\documentclass[letterpaper,10pt,twocolumn,DIV=15]{scrartcl}
\usepackage[T1]{fontenc}
\usepackage[utf8]{inputenc}
\usepackage[font={small}]{caption}
\usepackage[bookmarks=false,hidelinks]{hyperref}
\usepackage{times}
\usepackage{amsmath,amssymb,amsthm}
\usepackage[affil-it]{authblk}
\usepackage{listings}
\usepackage[round]{natbib}
\usepackage{microtype}
\usepackage[shortlabels]{enumitem}

\setkomafont{section}{\Large}
\setkomafont{subsection}{\large}
\addtokomafont{disposition}{\rmfamily}

\usepackage{graphicx}

\date{}

\usepackage[font={small}]{subcaption}
\usepackage{xspace}

\usepackage{xstring}

\usepackage{amsmath,amssymb}

\usepackage[ruled,vlined,linesnumbered]{algorithm2e}

\SetCommentSty{mycommentfont}
\DontPrintSemicolon
\SetKwComment{mycomment}{}{}

\DeclareMathOperator*{\argmin}{arg\,min}

\newcommand{\giv}[1]{\underline{#1}}

\renewcommand{\mid}{\,\vert\,}

\newcommand{\Expb}[2]{\mathbb{E}_{#2}\left[{#1}\right]}

\newcommand{\y}{\mathbf{y}}
\newcommand{\ysim}{\y}
\newcommand{\ygiv}{\giv{\y}}

\newcommand{\weights}{w}
\newcommand{\metric}{D}
\newcommand{\param}{\boldsymbol{\theta}}

\newcommand{\parambest}{\boldsymbol{\theta}_\star}
\newcommand{\N}[1]{\mathcal{N}\left(#1\right)}
\newcommand{\MC}{\mathcal{M}}
\newcommand{\pvalue}{\rho}
\newcommand{\dispersion}{d}

\newcommand{\illspec}{misspecified\xspace}

\newcommand{\itmc}{\textsc{Itmc}\xspace}

\graphicspath{{./figures/}}

\title{How consistent is my model with the data?\\ Information-Theoretic Model Check} 

\author{Andreas Svensson\thanks{\url{andreas.svensson@it.uu.se}}} 
\author{Dave Zachariah\thanks{\url{dave.zachariah@it.uu.se}}} 
\author{Thomas B. Sch\"on\thanks{\url{thomas.schon@it.uu.se}}} 
\affil{Department of Information Technology, Uppsala University}

\begin{document}

\maketitle

\begin{abstract}                % Abstract of not more than 250 words.
The choice of model class is fundamental in statistical learning and system identification, no matter whether the class is derived from physical principles or is a generic black-box. We develop a method to evaluate the specified model class by assessing its capability of reproducing data that is similar to the observed data record. This model check is based on the information-theoretic properties of models viewed as data generators and is applicable to e.g. sequential data and nonlinear dynamical models. The method can be understood as a specific two-sided posterior predictive test. We apply the information-theoretic model check to both synthetic and real data and compare it with a classical whiteness test.
\end{abstract}

\section{Introduction}

Parametric statistical inference often begins with the choice of a model class which is used to describe an unknown data-generating process. In system identification and sequential data analysis, we obtain a sequence of dependent samples from this  process. A classical problem has been to assess whether the unknown process is contained in the proposed model class, usually relying on large-sample results \citep{White1982_mlemisspecified}. In many real-world applications, however, we only have a limited data record and we expect the model class to be misspecified in some respect. A more relevant question would then be: \emph{how consistent is the model class with the observed data?}

A classical means of assessing a model is through its residuals or prediction errors. E.g. for linear dynamic models, one can check whether their prediction errors constitute a white noise process, cf. \citet{LB:78} and \citet[ch.~11]{SoderstromStoica1988_system}. In such cases, the errors reflect an irreducible component of the data-generating process and the model class is deemed consistent with the data. The whiteness statistic, however, only assesses the maximum likelihood model in the proposed model class, and it is not trivial to extend the whiteness statistics to more general models, such as, e.g., nonlinear state-space models. 

In this paper, we develop a method to quantify the consistency of more general model classes to observed data. It follows the principle that `if the model fits, then replicated data generated under the model should look similar to observed data' \citep[p. 143]{GCS+:14}. The proposed measure targets those models in the class that provide the best approximation of the data-generating process and it corresponds to a p-value in a Portmanteau test, in that no alternative model classes have to be specified. The method is based on an  information-theoretic formulation that quantifies the `similarity' between observed data and data that is generated by the model. For this reason, we will refer to our method as an information-theoretic model check (\itmc). From a Bayesian perspective, the \itmc can be understood as a posterior predictive test (e.g., \citealt{R:84}, \citealt{GMS:96}, \cite[Section 6.3]{GCS+:14}) using a test statistic with two tail events. We will, however, motivate our  contribution from a more general perspective centered on the best models in a specified model class.

For models with latent variables (such as state-space models) another method was recently proposed by \citet{SMW:17}, with the idea of assessing the properties of the posterior distribution for the latent variables (i.e., the state smoothing distribution for state-space models). We will compare the proposed \itmc to this, as well as to the Ljung-Box method.

\emph{Notation:} For notational convenience, we denote a random sequence as $\y \equiv \{ \y_1, \y_2, \dots, \y_T \}$ and distinguish it from the observed data $\ygiv \equiv \{ \ygiv_1, \ygiv_2, \dots, \ygiv_T \}$ in the same space. We denote the number of data samples in the sequence as $T$, and make no assumptions about independence among the samples. Possible known exogenous input signals are omitted from the notation.

\section{Information-theoretic model check}
Let $p_0(\y)$ denote an unknown data-generating process, from which we have obtained a data sequence $\ygiv$. We are interested in a mathematical description of $p_0(\y)$, using the knowledge provided to us via $\ygiv$. Such inference begins by specifying a class of models, each of which corresponds to a possible data-generating distribution $p(\y|\param)$ indexed by $\param$. We denote such a parametric model class
\begin{equation}
\MC = \big\{ p(\y | \param ) \: : \: \param \in \Theta \big\},
\end{equation}
where $\Theta\subset\mathbb{R}^d$. Our aim is to assess the consistency of this class with respect to the observed data $\ygiv$. Specifically, we will assess whether the models in $\MC$ that best approximate $p_0(\y)$ could generate sequences similar to the observed sequence $\ygiv$.

Using the Kullback-Leibler divergence \citep{Kullback1997_information} as our metric, we index the minimum divergence models by
\begin{equation}
\parambest \: \in \: \argmin_{\param} \; \underbrace{\Expb{ \ln p_0(\y) - \ln p(\y | \param)  }{0}}_{\text{model divergence}},
\label{eq:mindivergencemodels}
\end{equation}
where the expectation is over $\y\sim p_0(\y)$. When the minimum divergence is 0, the model class is well-specified. That is, $p(\y | \parambest)$ and $p_0(\y)$ generate statistically indistinguishable sequences $\y$. The model divergence, however, depends on the unknown data-generating process $p_0(\y)$ and can therefore not be computed directly. Our starting point therefore is to assess how probable it is that the best models in the class, $p(\y | \parambest)$, could generate sequences that are similar to $\ygiv$ observed from $p_0(\y)$.

\subsection{Assessment of model as a data-generator}

\begin{figure}[ht]
	\centering
	\includegraphics{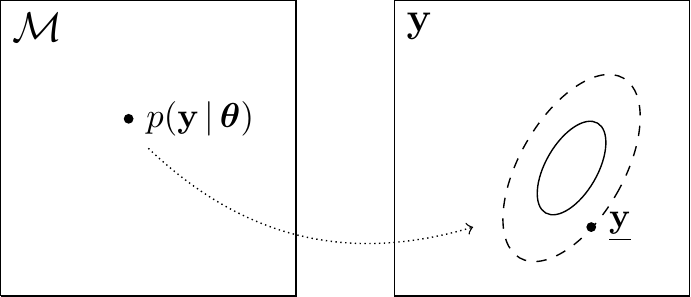}
	%\tikzsetnextfilename{fig_1}
	%\begin{tikzpicture}
	%	% y square
	%	\coordinate (yll) at (4, 0);
	%	\coordinate [label={below right:\large$\y$}] (yul) at (4, 3);
	%	\coordinate (yur) at (7, 3);
	%	\coordinate (ylr) at (7, 0);
	%	\coordinate (yc) at (5.8,1.3);
	%	\node[circle,fill=black,inner sep=1pt,label=right:\textbf{$\ygiv$}] (y) at (6.0,0.7) {};
	%	
	%	\draw[dashed,rotate around={-30:(yc)}] (yc) ellipse (15pt and 30pt);  
	%	\draw[rotate around={-30:(yc)}] (yc) ellipse (7.5pt and 15pt);  
	%% 	\draw[rotate around={-30:(yc)}] (yc) ellipse (3pt and 6pt);  	
	%	
	%
	%	% M square
	%	\coordinate (mll) at (0, 0);
	%	\coordinate [label={below right:\large$\MC$}] (mul) at (0, 3);
	%	\coordinate (mur) at (3, 3);
	%	\coordinate (mlr) at (3, 0);
	%	\node[circle,fill=black,inner sep=1pt,label=right:{$p(\y\mid\param)$}] (py) at (1.3,1.8) {};
	%% 	\node[circle,fill=black,inner sep=1pt,label=right:{$p_0(\y)$}] (po) at (6,-0.3) {};
	%	
	%	% squares
	%	\draw (yll)--(yul)--(yur)--(ylr)--(yll);
	%	\draw (mll)--(mul)--(mur)--(mlr)--(mll);
	%	
	%	% dotted lines
	% 	\draw[bend right,densely dotted,->] ($(py)+(0.2,-0.3)$) to ($(yc)+(-1,-.6)$);
	%% 	\draw[dashed] ($(po)+(-.1,0)$) parabola ($(y)+(0.05,-0.05)$);
	%\end{tikzpicture}
	\caption{Every fix model $p(\y | \param)$ in the model class $\MC$ (left) generates hypothetical sequences $\y$ (right). For any given $\y$, the surprisal $\metric(\y, \param)$ may be greater or less than $D(\ygiv,\param)$. The dashed contour represent the set \eqref{eq:surprisalset} with equal levels of surprisal as $\ygiv$, whereas the solid contour represents a set of sequences $\y$ with $\metric(\y, \param)\leq \metric(\ygiv, \param)$. A model that tends to generate sequences $\y$ with similar levels of surprisal as that of $\ygiv$, is deemed to be highly consistent with the data. This principle motivates \eqref{eq:pvalueclassic}.}
	\label{fig:1}
\end{figure}

Consider a specific model in $\MC$, which generates a random sequence:
\begin{equation*}
\y \sim p(\y | \param).
\end{equation*}
We expect certain types of generated sequences to be more probable than others \citep{Cover&Thomas2012_elements,Gray2011_entropy} and this information-theoretic principle enables us to assess whether the model could generate $\y$ similar to $\ygiv$. We use self-information or the `surprisal' of observing $\y$,
\begin{equation*}
\metric(\y, \param) \triangleq -\ln p(\y | \param),
\end{equation*}
as a test statistic. Since $\y$ is random, the surprisal is a random variable and its expected value is the entropy of the model. Based on this quantity, we define a set of sequences that are equivalent to $\ygiv$,
\begin{equation}
\big \{ \y : \metric(\y, \param) = \metric(\ygiv, \param) \big\},
\label{eq:surprisalset}
\end{equation}
i.e., a set where each $\y$ has the same surprisal. This is illustrated by the dashed contour in Figure~\ref{fig:1}.

Now consider two tail events: $\y$ has either higher or lower surprisal than $\ygiv$. When the probabilities of both events are balanced, the surprisal of $\ygiv$ is within the typical range of surprisals for sequences generated by model $p(\y | \param)$. We can define a p-value for the two tail events:
\begin{equation}
\begin{split}
\pvalue(\ygiv, \param) = 2 \min\big\{ & \Pr\{ \metric(\y,\param) \geq \metric(\ygiv,\param) \}, \\
& \Pr\{ \metric(\y,\param) \leq \metric(\ygiv,\param) \} \big\}.
\end{split}
\label{eq:pvalueclassic}
\end{equation}
Thus when $\pvalue(\ygiv, \param) \in [0,1]$ is very low, $p(\y | \param)$ tends to generate sequences $\y$ with very different levels of surprisal when compared to the actual observations $\ygiv$.

\emph{Remark 1:} The model divergence in \eqref{eq:mindivergencemodels} is equivalent to the difference in surprisal of $\y$ under the model $p(\y| \param)$ and the data generating process, on average.

\emph{Remark 2:} For models generating $\y$ with i.i.d. blocks, the sets of sequences with equal surprisal are related to the notion of typical sets, cf. \citet{Cover&Thomas2012_elements}.

\emph{Remark 3:} The p-value for the Ljung-Box method is based on a one tail event, unlike \eqref{eq:pvalueclassic}.

\subsection{Averaging around minimum divergence models}

Ideally, the minimum divergence models \eqref{eq:mindivergencemodels} would be evaluated using \eqref{eq:pvalueclassic}. That is, evaluate $\pvalue(\ygiv, \parambest)$ where $\parambest$ is unknown. In lieu of $\parambest$, we define the averaged p-value,
\begin{equation}
\begin{split}
\boxed{\pvalue_\star(\ygiv) = \int \pvalue(\ygiv, \param) \weights(\param | \ygiv) \: d\param,}
\end{split}
\label{eq:itmc}
\end{equation}
where the weights $\weights(\param | \ygiv) \geq 0$ are high for models in the neighbourhood of $\parambest$. The weights are normalized and defined as follows
\begin{equation}
\weights(\param | \ygiv ) \triangleq \frac{\weights_0(\param) p(\ygiv|\param)}{\int \weights_0(\param) p(\ygiv|\param) d\param},
\label{eq:weights}
\end{equation}
which reflects uncertainty about where $\parambest$ is located in the parameter space \citep{Casella&Berger2002_statistical,BissiriEtAl2012_converting}. The initial weights $w_0(\param)$ are defined using one of the following approaches when $\parambest$ is unique:
\begin{itemize}
	\item \emph{Fisherian:} By setting $w_0(\param) \propto $ constant, the maximum weight \eqref{eq:weights} is located at the maximum likelihood estimate which converges to $\parambest$ as $T \rightarrow \infty$ under standard regularity conditions \citep{LC:79}.
	
	\item \emph{Bayesian:} Alternatively, $\weights_0(\param)$ can be chosen to describe prior assumptions about the location of $\parambest$. Then \eqref{eq:weights} represents the posterior distribution of the parameters. When the sequence consists of blocks that are i.i.d., the weights \eqref{eq:weights} concentrate at $\parambest$ as $T \rightarrow \infty$ \citep{B:66}.
\end{itemize}

We refer to~\eqref{eq:itmc}, with either of the choices for $\weights_0(\param)$, as the information-theoretic model check (\itmc). When $\pvalue_\star(\ygiv)$ is close to zero, it is highly improbable that the minimum divergence model can generate sequences that are similar to $\ygiv$. In this case, the specified model class $\MC$ is highly inconsistent with the data.

As the weights concentrate around $\parambest$, the dispersion of $\pvalue(\ygiv, \param)$ in this neighbourhood, i.e.,
\begin{equation}
\begin{split}
\dispersion_\star(\ygiv) = \left[\int \big(\pvalue(\ygiv, \param) - \pvalue_\star(\ygiv)\big)^2 \weights(\param | \ygiv) \: d\param \right]^{1/2},
\end{split}\label{eq:dispersion}
\end{equation}
is reduced. Thus an interval $\pvalue_\star(\ygiv) \pm 2 \dispersion_\star(\ygiv)$ quantifies the confidence in the consistency assessment of $\MC$.

\emph{Remark 1:} In the context of single data or independent data,  \citet{Box1980_sampling} considered a statistic for one-tail events using a model averaged over the entire model class, i.e., $p(\y) = \int p(\y|\param) \weights_0(\param) d \param$. This approach does not target $\parambest$ of interest here, cf. \citet[Section 5.3]{R:84} for a more elaborate discussion.\footnote{In the non-trivial case of a state-space model, which we will illustrate in Section~\ref{sec:ex:WT}, we actually consider the `posterior' for the unknown parameters $\param$ but the `prior' for the equally unknown latent states. This is a natural choice for state-space models, and can yet again be motivated from the discussion by \citet{R:84} on what should be considered as a replication for the problem at hand.} 

\emph{Remark 2:} \citet{GMS:96} defined a Bayesian posterior p-value for general statistics with one-tail events. The measure \eqref{eq:itmc} is similar but uses a specific information-theoretic test statistic with two-tail events and also quantifies the assessment via \eqref{eq:dispersion}.

\subsection{Implementation}

To compute $\pvalue_\star(\ygiv)$ in \eqref{eq:itmc}, several integrals which are not readily available in closed form have to be evaluated. These can be approximated numerically using a straightforward Monte Carlo implementation as outlined in Algorithm~\ref{alg:impl}. With $N$ being the number of $\param$-samples from $\weights(\param\mid\ygiv)$ and $M$ the number of $\y$-samples, for each $\param$-sample, from $p(\ysim\mid\param)$, it has a computational complexity in the order of $NM$. We have found a satisfactory performance of this implementation with only moderate values for $N$ and $M$.

Here we assume that new data can be simulated from $p(\ysim\mid\param)$, and that samples can be drawn from $\weights(\param\mid\ygiv)$. The former assumption is natural for many model classes, and methods such as importance sampling or Markov chain Monte Carlo can be used for the latter case if needed (as will be done in Section~\ref{sec:ex:WT}). We further assume that $p(\y\mid\param)$ is available also for point-wise evaluation, which is the case for many (but not all) models. For general state-space models (as in Section~\ref{sec:ex:WT}) we can only estimate this using, e.g., a particle filter \citep{DoucetJ:2011}.

\begin{algorithm}[ht]
	Construct $\weights(\param\mid\ygiv)$\;
	Draw $N$ samples $\param^{(i)}\sim \weights(\param\mid\ygiv)$\;
	\For{$i = 1, \dots, N$}{
		Simulate $M$ trajectories $\ysim^{(j)}\sim p(\ysim\mid\param^{(i)})$\;
		Compute $D^{(i,j)} = -\ln p(\ysim^{(j)}\mid\param^{(i)})$ for $j=1, \dots, M$\;
		Compute $\giv{D}^{(i)} = -\ln p(\ygiv\mid\param^{(i)})$\;
		Set $\widehat{\pvalue}^{(i)} = 2\min\big\{\frac{1}{M}\sum_{j=1}^M(D^{(i,j)}\geq\giv{D}^{(i)}),\linebreak ~ \hspace{7.15em} \frac{1}{M}\sum_{j=1}^M(D^{(i,j)}\leq\giv{D}^{(i)})\big\}$
	}
	Set $\widehat{\pvalue}_\star(\ygiv) = \frac{1}{N}\sum_{i=1}^N\widehat{\pvalue}^{(i)}$\;
	\caption{Monte Carlo implementation}
	\label{alg:impl}
\end{algorithm}

\section{Illustrations and experiments}

We will first consider a series of examples with synthetic data. The behavior of the model check is illustrated and then compared to the well-established Ljung-Box method. Finally the method is applied to a non-trivial real-data system identification problem. All source code is available via GitHub\footnote{https://github.com/saerdna-se/itmc}.

\subsection{Synthetic data: illustration of method}

\begin{figure}[t]
	\begin{subfigure}{\linewidth}
		\centering
		\includegraphics{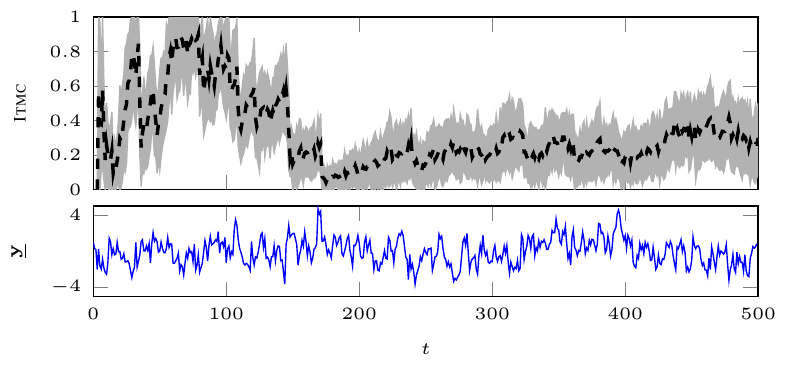}
		%		\tikzsetnextfilename{ex_disp_well}
		%		\begin{tikzpicture}
		%    		\begin{axis}[at={(0,0.13\linewidth)},width=\linewidth,height=.4\linewidth,xmin=0,xmax=500,ylabel=\itmc,ymin=0,ymax=1,xtick distance={100},xticklabels={}]
		%        		\addplot[name path=U,mark=none,color=black!30!white] table [x index=0,y expr={min(\thisrowno{1}+2*sqrt(\thisrowno{2}),1)}] {Example4_dispersion/res_0b.csv};
		%        		\addplot[name path=L,mark=none,color=black!30!white] table [x index=0,y expr={max(\thisrowno{1}-2*sqrt(\thisrowno{2}),0)}] {Example4_dispersion/res_0b.csv};
		%        		\addplot[black!30!white] fill between[of=U and L];
		%        		\addplot[mark=none,color=black,line width=1pt,dashed] table [x index=0,y index=1] {Example4_dispersion/res_0b.csv};
		%    		\end{axis}
		%    		    \begin{axis}[at={(0,0)},width=\linewidth,height=.3\linewidth,xmin=0,xmax=500,xlabel=$t$,ylabel=$\ygiv$,ymax=5,ymin=-5,ytick={-4,4},xtick distance={100}]
		%    		\addplot[mark=none,color=blue] table [x expr=\coordindex, y index=0] {Example4_dispersion/y_0b.csv};
		%    		\end{axis}
		%		\end{tikzpicture}
		\caption{Case i (well-specified). As $t$ grows and data accumulates, the value of $\pvalue_\star(\ygiv)$ changes, but the overall conclusion from the \itmc is that the specified model class is consistent with $\ygiv$ throughout.}
	\end{subfigure}
	
	\vspace{2em}
	
	\begin{subfigure}{\linewidth}
		\centering
		\includegraphics{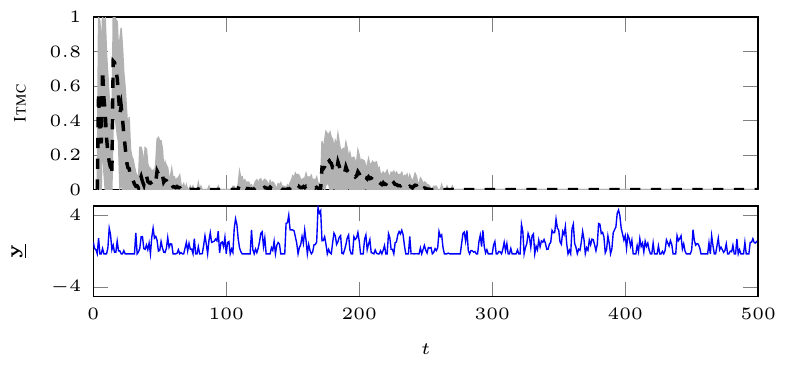}
		%		\tikzsetnextfilename{ex_disp_ill}
		%		\begin{tikzpicture}
		%    		\begin{axis}[at={(0,0.13\linewidth)},width=\linewidth,height=.4\linewidth,xmin=0,xmax=500,ylabel=\itmc,ymin=0,ymax=1,xtick distance={100},xticklabels={}]
		%        		\addplot[name path=U,mark=none,color=black!30!white] table [x index=0,y expr={min(\thisrowno{1}+2*sqrt(\thisrowno{2}),1)}] {Example4_dispersion/res_1b.csv};
		%        		\addplot[name path=L,mark=none,color=black!30!white] table [x index=0,y expr={max(\thisrowno{1}-2*sqrt(\thisrowno{2}),0)}] {Example4_dispersion/res_1b.csv};
		%        		\addplot[black!30!white] fill between[of=U and L];
		%        		\addplot[mark=none,color=black,line width=1pt,dashed] table [x index=0,y index=1] {Example4_dispersion/res_1b.csv};
		%    		\end{axis}
		%    		    \begin{axis}[at={(0,0)},width=\linewidth,height=.3\linewidth,xmin=0,xmax=500,xlabel=$t$,ylabel=$\ygiv$,ymax=5,ymin=-5,ytick={-4,4},xtick distance={100}]
		%    		\addplot[mark=none,color=blue] table [x expr=\coordindex, y index=0] {Example4_dispersion/y_1b.csv};
		%    		\end{axis}
		%		\end{tikzpicture}
		\caption{Case ii (\illspec). Here $\pvalue_\star(\ygiv)$ drops to zero and thus with high confidence the \itmc finds the specified model class to inconsistent with the data. Note that during the long period $80 \leq t \leq 200$, the saturation of $\y_t$ is largely inactive, which makes the model-data mismatch impossible to detect. By contrast, the notable saturated behavior at $t\approx 250$ is picked up by \itmc.}
	\end{subfigure}
	\caption{The \itmc illustrated by $\pvalue_\star(\ygiv)$ \eqref{eq:itmc} in black and its dispersion ($\pm 2 \dispersion_\star$) in gray, computed cumulatively for one data set from Case i (a) and Case ii (b), respectively. The lower panels in each subfigure illustrate the observed sequences $\ygiv$.}\label{fig:ex:disp}
\end{figure}

In the interest of clarity, we consider the simple class of AR(1)-models,
\begin{equation}
\MC = \Big\{ p(\y|\param) \: : \:  y_t = \param y_{t-1} + e_t, ~~ e_t\sim\N{0,1} \Big\},
\label{eq:ex:mc}
\end{equation}
throughout this section, with the auto-regression coefficient $\param$ as the only unknown parameter. For sake of generality, but of lesser significance herein, we consider initial weights $\weights_0(\param)$ as a zero-mean Gaussian with unit variance to reflect our initial beliefs about the location of the minimum divergence model in the parameter space. The following two  cases are studied:
\begin{itemize}
	\item[]\textbf{Case i}: The data generated by~\eqref{eq:ex:mc} with $\param=0.7$.
	\item[]\textbf{Case ii}: The data generated by a saturated version of~\eqref{eq:ex:mc}, namely
	\begin{equation}
	y_t = \left(0.7 y_{t-1} + e_t\right)\vee-0.3, ~~ e_t\sim\N{0,1}.\label{eq:ex:ill2}
	\end{equation}
\end{itemize}
The model class~\eqref{eq:ex:mc} is thus well-specified for Case i, but \illspec for Case ii. For illustration, we compute the \itmc (and its dispersion, using Algorithm~\ref{alg:impl} with $N=20$ and $M=50$) as data samples $\ygiv_t$ are added sequentially up to $T=500$. The results are plotted in Figure~\ref{fig:ex:disp}, and we see that the \itmc indicates that the model class is not consistent with the data in Case ii.

\subsection{Synthetic data: evaluation and comparison}

In this section, we will evaluate the performance of the proposed \itmc by repeating all experiments 100 times each. We also compare all results to the well-established Ljung-Box method \citep{LB:78}, which builds on the test statistic
\begin{equation}
Q = T(T+2)\sum_{k=1}^h\frac{\widehat{r}_k^2}{T-k},\label{eq:lb}
\end{equation}
where $\widehat{r}_k$ is the lag $k$ sample autocorrelation of the prediction errors of an estimated model typically using the maximum likelihood method. If the minimum divergence model yields white residuals (null hypothesis), $Q$ follows a $\chi^2_{h-d}$ distribution. From this one can construct a $p$-value to measure the evidence against the null hypothesis. A similar method is formulated in  \citet[ch.~11]{SoderstromStoica1988_system}. Unlike \itmc, the Ljung-Box method concentrates on a single estimated model in a linear model class.

\begin{figure}[t]
	\begin{subfigure}{\linewidth}
		\centering
		\includegraphics{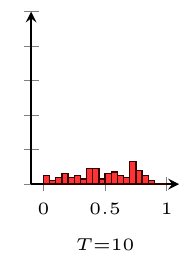}
		\includegraphics{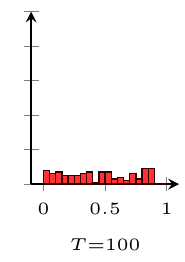}
		\includegraphics{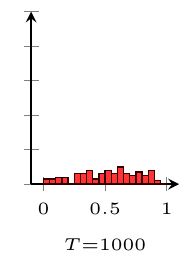}
		\includegraphics{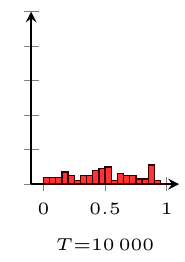}
		%	\addhistogram{Example5/met_10.csv}{$T\text{$=$}10$}
		%	\addhistogram{Example5/met_100.csv}{$T\text{$=$}100$}
		%	\addhistogram{Example5/met_1000.csv}{$T\text{$=$}1000$}
		%	\addhistogram{Example5/met_10000.csv}{$T\text{$=$}10\,000$}
		\caption{\itmc.}
	\end{subfigure}
	\begin{subfigure}{\linewidth}
		\centering
		\includegraphics{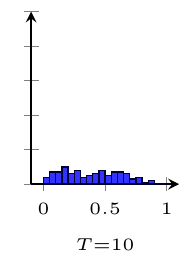}
		\includegraphics{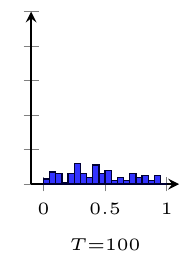}
		\includegraphics{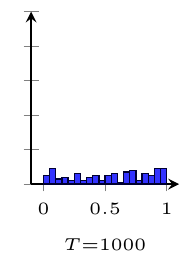}
		\includegraphics{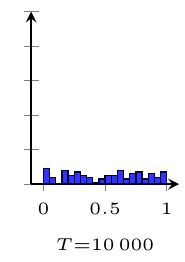}
		%	\addhistograml{Example5/lb_10.csv}{$T\text{$=$}10$}
		%	\addhistograml{Example5/lb_100.csv}{$T\text{$=$}100$}
		%	\addhistograml{Example5/lb_1000.csv}{$T\text{$=$}1000$}
		%	\addhistograml{Example5/lb_10000.csv}{$T\text{$=$}10\,000$}
		\caption{The Ljung-Box method.}
	\end{subfigure}
	\caption{Case i (well-specified). Statistical behavior of the \itmc (a) and the Ljung-Box (b) method when the data is generated by a model in the model class $\MC$ given in~\eqref{eq:ex:mc}, for various values of $T$. Each histogram reports results from $100$ individual realizations.}\label{fig:ex:well}
\end{figure}

We start by repeating the well-specified Case i from the previous section. For each of the batch sizes $T=10, 100, 1000$ and $10\,000$, we simulate $100$ such data sets, and apply the \itmc  and the Ljung-Box method (with $h\approx\log T$). The results are shown as histograms in Figure~\ref{fig:ex:well}. For both methods the histograms are nearly uniform, similar to classical p-values which have a uniform distribution under the null hypothesis. 

\begin{figure}[t]
	\begin{subfigure}{\linewidth}
		\centering
		\includegraphics{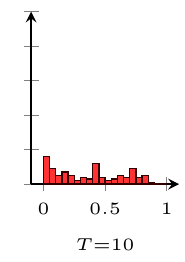}
		\includegraphics{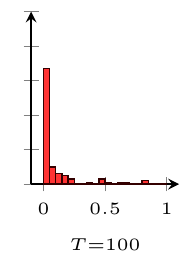}
		\includegraphics{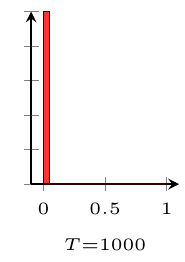}
		\includegraphics{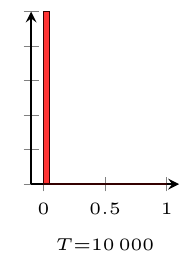}
		%	\addhistogram{Example4/met_1_10.csv}{$T\text{$=$}10$}
		%	\addhistogram{Example4/met_1_100.csv}{$T\text{$=$}100$}
		%	\addhistogram{Example4/met_1_1000.csv}{$T\text{$=$}1000$}
		%	\addhistogram{Example4/met_1_10000.csv}{$T\text{$=$}10\,000$}
		\caption{\itmc.}
	\end{subfigure}
	\begin{subfigure}{\linewidth}
		\centering
		\includegraphics{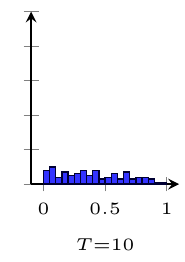}
		\includegraphics{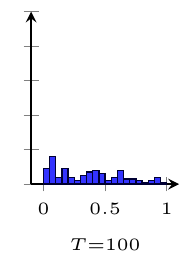}
		\includegraphics{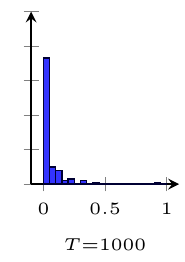}
		\includegraphics{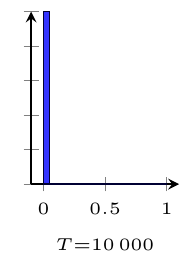}
		%	\addhistograml{Example4/lb_1_10.csv}{$T\text{$=$}10$}
		%	\addhistograml{Example4/lb_1_100.csv}{$T\text{$=$}100$}
		%	\addhistograml{Example4/lb_1_1000.csv}{$T\text{$=$}1000$}
		%	\addhistograml{Example4/lb_1_10000.csv}{$T\text{$=$}10\,000$}
		\caption{The Ljung-Box method.}
	\end{subfigure}
	\caption{Case ii (\illspec). Both methods detect an inconsistent model class with respect to the data, but at different data lengths $T$.}\label{fig:ex:ill2}
\end{figure}

\begin{figure}[t]
	\begin{subfigure}{\linewidth}
		\centering
		\includegraphics{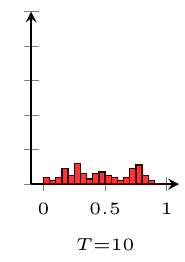}
		\includegraphics{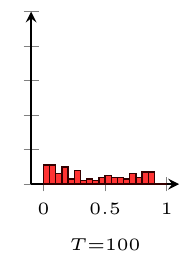}
		\includegraphics{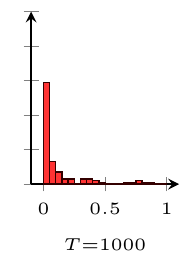}
		\includegraphics{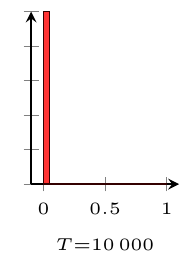}
		%	\addhistogram{Example4/met_0_10.csv}{$T\text{$=$}10$}
		%	\addhistogram{Example4/met_0_100.csv}{$T\text{$=$}100$}
		%	\addhistogram{Example4/met_0_1000.csv}{$T\text{$=$}1000$}
		% 	\addhistogram{Example4/met_0_10000.csv}{$T\text{$=$}10\,000$}
		\caption{\itmc.}
	\end{subfigure}
	\begin{subfigure}{\linewidth}
		\centering
		\includegraphics{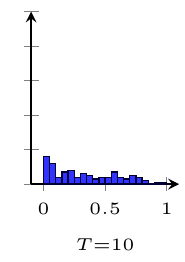}
		\includegraphics{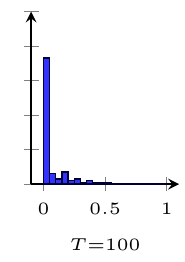}
		\includegraphics{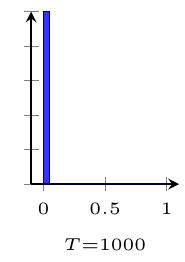}
		\includegraphics{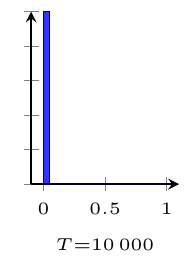}
		%	\addhistograml{Example4/lb_0_10.csv}{$T\text{$=$}10$}
		%	\addhistograml{Example4/lb_0_100.csv}{$T\text{$=$}100$}
		%	\addhistograml{Example4/lb_0_1000.csv}{$T\text{$=$}1000$}
		%	\addhistograml{Example4/lb_0_10000.csv}{$T\text{$=$}10\,000$}
		\caption{The Ljung-Box method.}
	\end{subfigure}
	\caption{Case iii (\illspec). The misspecification is less severe than in Case ii (Figure~\ref{fig:ex:ill2}).}\label{fig:ex:ill1}
\end{figure}

We now consider the \illspec Case ii, together with another \illspec \textbf{Case iii}, where the data is generated by
\begin{equation}
y_t = -0.3 y_{t-1} + 0.5y_{t-2}+ e_t, ~~ e_t\sim\N{0,1},\label{eq:ex:ill1}
\end{equation}
i.e., an AR(2)-model. Note that in case iii, the model misspecification is less severe than in Case ii: The best AR(1) model is more likely to generate data similar to \eqref{eq:ex:ill1} than to \eqref{eq:ex:ill2}. As in the well-specified case, we apply the \itmc along with the Ljung-Box method, and report the results for Case ii and iii in Figure~\ref{fig:ex:ill2} and \ref{fig:ex:ill1}, respectively.

The results indicate that when $T$ becomes very large,  both the \itmc and the Ljung-Box method ultimately find $\MC$ to be inconsistent with the data in Case ii and iii. For \itmc there is a noticeable difference between both cases, in that it takes fewer samples to find $\MC$ to be inconsistent in the more severely misspecified Case ii. For the Ljung-Box method, which targets a different property of the model class, the behavior for Case ii and iii is reversed.

\begin{figure}[t]
	\begin{subfigure}{.49\linewidth}
		\centering
		\includegraphics{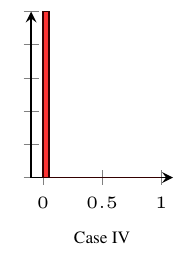}
		\includegraphics{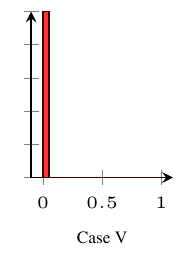}
		%	\addhistograms{Example3/met_2.csv}{Case IV}
		%	\addhistograms{Example3/met_3.csv}{Case V}
		\caption{\itmc.}
	\end{subfigure}
	\begin{subfigure}{.49\linewidth}
		\centering
		\includegraphics{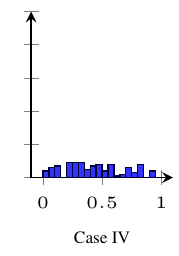}
		\includegraphics{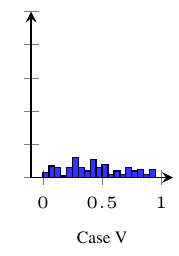}
		%	\addhistogramls{Example3/lb_2_100.csv}{Case IV}
		%	\addhistogramls{Example3/lb_3_100.csv}{Case V}
		\caption{The Ljung-Box method.}
	\end{subfigure}
	\caption{Case iv and v, $T = 100$. While the Ljung-Box method (b) behaves as in the well-specified case (Figure~\ref{fig:ex:well}), the \itmc (a) clearly indicates the \illspec variances.}\label{fig:ex:ill3}
	\vspace{-1em}
\end{figure}

For simplicity, we have so far assumed the noise variances to be known. In most applications, however, this is not the case. To illustrate that the \itmc---in contrast with the Ljung-Box method---also indicates incorrectly specified variances, we consider data sets with $T=100$ generated by
\begin{equation}
y_t = 0.7 y_{t-1} + e_t, ~~ e_t\sim\N{0,0.1},
\end{equation}
still (erroneously) assuming~\eqref{eq:ex:mc} as the model class (\textbf{Case iv}), as well as the opposite (\textbf{Case v}) with data generated with noise variance $1$ but the model class assumes a variance of only $0.1$. We report these result in Figure~\ref{fig:ex:ill3}. As expected, the Ljung-Box method does not register any problems with $\MC$ since it is constructed as a pure whiteness test, in contrast to \itmc. It is however more natural to assume that the noise variance is one of the unknown parameters $\param$ in the model class $\MC$. This will indeed be the case in the next example.

%%%%%%%%%%%%%%%%%%%%%%%%%%%%%%%%%%%%%%%%%%%%%%%%%%%%%
\subsection{Real data: cascaded water tank modeling}\label{sec:ex:WT}
\begin{figure}[ht]
	\centering
	\includegraphics[width=.5\linewidth]{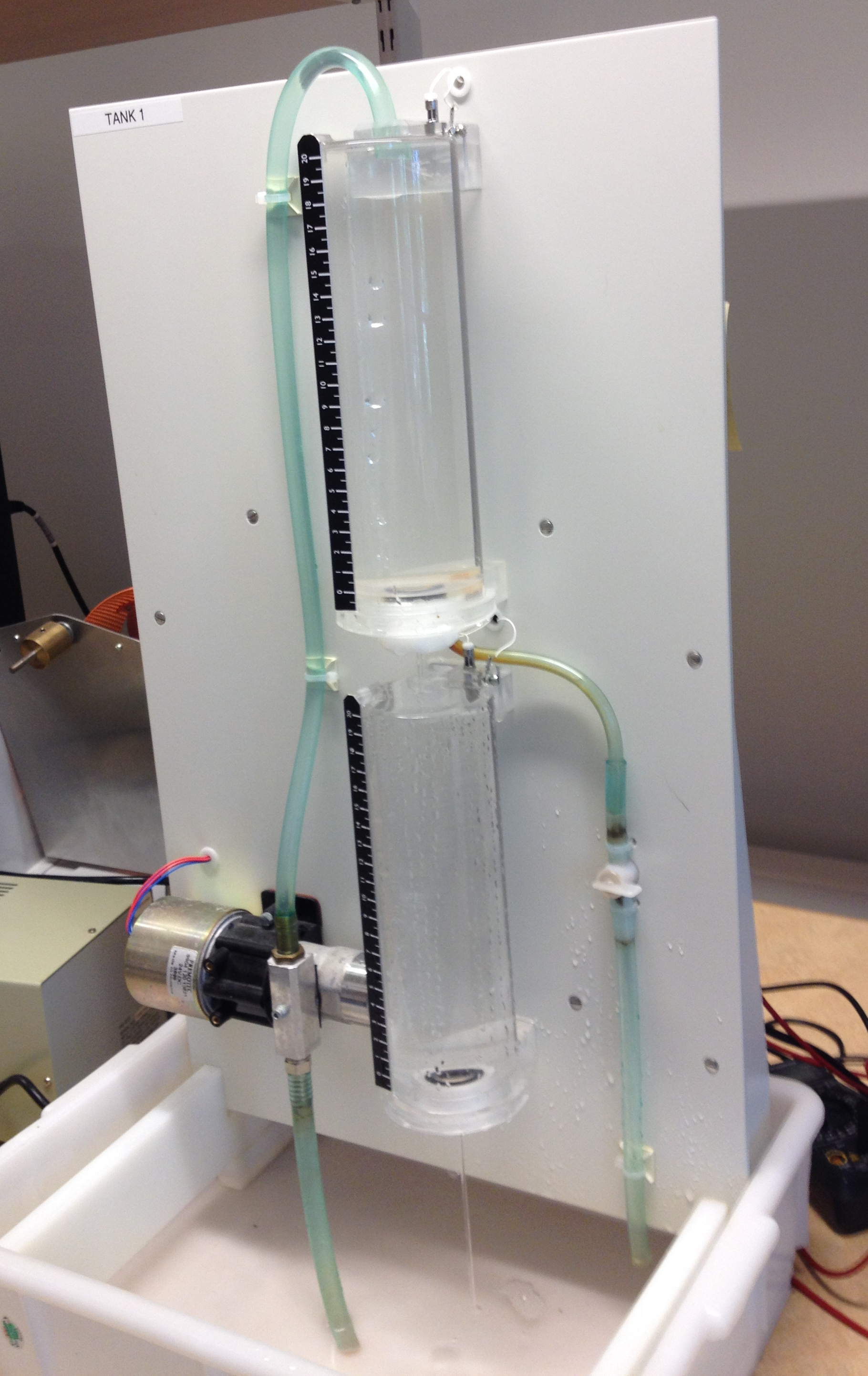}
	\caption{Cascaded water tanks. We consider the problem of modelling the behavior from input signal (voltage over the pump generating the inflow to the upper tank) to the output signal (water level in the lower tank). Thanks to Maarten Schoukens for the picture in Figure~\ref{fig:exWT}. Photo courtesy to Maarten Schoukens.}
	\label{fig:wt:pic}\vspace{-1em}
\end{figure}

Cascaded water tanks, as shown in Figure~\ref{fig:wt:pic}, are well-studied in the automatic control and system identification community. A commonly used first-principles state-space model to describe the system is
%\begin{subequations}
	\begin{align}
	\dot{x}_1(t) &= - k_1\sqrt{x_1(t)} + k_2 u(t) + w_1(t),\nonumber\\
	\dot{x}_2(t) &= k_1\sqrt{x_1(t)} - k_3\sqrt{x_3(t)} + w_2(t),\nonumber\\
	y(t) &= x_2(t),\label{eq:WT:org}
	\end{align}
%\end{subequations}
where $k_1$, $k_2$, and $k_3$ have a physical interpretation, and has to be identified along with the variance of the white noise $w_1(t)$ and $w_2(t)$.
At the nonlinear system identification workshop in Brussels 2016, it was pointed out by \citet{HRC+:16} that this model relies on an improbable assumption of laminar flow. A better model of the system, grounded in fluid dynamics, would be
%\begin{subequations}
{\small
	\begin{align}
	\dot{x}_1(t) &= - k_1\sqrt{x_1(t)} - k_4x_1(t) + k_2 u(t) + w_1(t),\nonumber\\
	\dot{x}_2(t) &= k_1\sqrt{x_1(t)} + k_4x_1(t) - k_3\sqrt{x_3(t)} - k_5x_1(t) + w_2(t),\nonumber\\
	y(t) &= x_2(t),\label{eq:WT:ext}
	\end{align}
}%\end{subequations}
where $k_4$ and $k_5$ have to be identified as well. Indeed, after parameter estimation, the predictive performance of~\eqref{eq:WT:ext} on test data is in general superior to the original model~\eqref{eq:WT:org}. To complicate the picture slightly, the data set provided by \citet{SN:17} also contained events with overflow, for which the model was extended as well, see \citet{HRC+:16}.

\begin{figure}[ht]
	\centering
	\begin{subfigure}{\linewidth}
		\centering
		\includegraphics{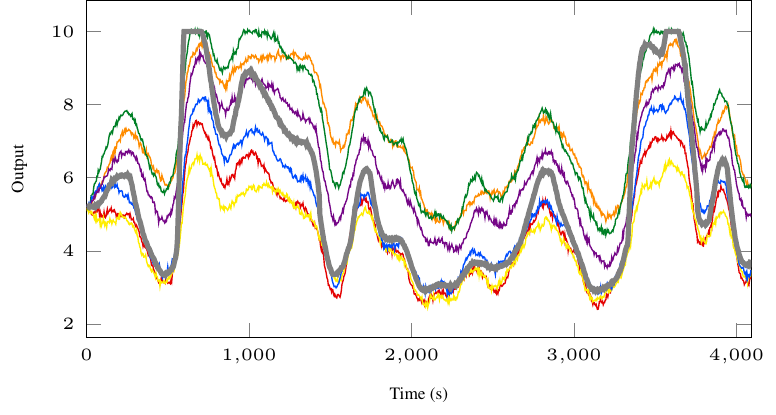}
		%		\tikzsetnextfilename{exWT1}
		%		\begin{tikzpicture}
		%		\begin{axis}[width=\linewidth,height=0.6\linewidth,xlabel=Time (s),ylabel=Output,xmin=0,xmax=4092,cycle list name=farger]
		%			\addplot table [col sep=comma,x expr=4*\coordindex,y index=1] {WT/y.csv};
		%			\addplot table [col sep=comma,x expr=4*\coordindex,y index=2] {WT/y.csv};
		%			\addplot table [col sep=comma,x expr=4*\coordindex,y index=3] {WT/y.csv};
		%			\addplot table [col sep=comma,x expr=4*\coordindex,y index=4] {WT/y.csv};
		%			\addplot table [col sep=comma,x expr=4*\coordindex,y index=5] {WT/y.csv};
		%			\addplot table [col sep=comma,x expr=4*\coordindex,y index=6] {WT/y.csv};
		%			\addplot+[line width=1.5pt] table [col sep=comma,x expr={4*\coordindex},y index=0] {WT/y.csv};
		%		\end{axis}
		%		\end{tikzpicture}
		\caption{The real data $\ygiv$ (gray) together with synthetic data (colored) from different models in the model class~\eqref{eq:WT:ext} (each colored data set is generated with randomly chosen physically reasonable parameter values). Intuitively, the \itmc assesses whether the gray is also likely to come from that model  class.}\label{fig:exWT:1}
	\end{subfigure}
	
	\vspace{2em}
	
	\begin{subfigure}{\linewidth}
		\centering
		\includegraphics{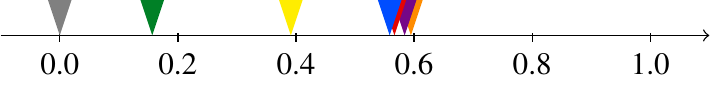}
		%		\tikzsetnextfilename{exWT2}
		%		\begin{tikzpicture}[scale=0.6]
		%		\draw[->] (-1,0) -- (11,0);
		%		%ticks
		%		\foreach \x in {0,2,4,6,8}
		%		\draw (\x,1pt) -- (\x,-3pt) node[anchor=north] {\small 0.\x};
		%		\draw (10,1pt) -- (10,-3pt) node[anchor=north] {\small 1.0};
		%		\fill[gray] (0,0)--(0+.2,.6)--(0-.2,.6);
		%		\fill[colo] (5.9467,0)--(5.9467+.2,.6)--(5.9467-.2,.6);
		%		\fill[colp] (5.84,0)--(5.84+.2,.6)--(5.84-.2,.6);
		%		\fill[coly] (3.9133,0)--(3.9133+.2,.6)--(3.9133-.2,.6);
		%		\fill[colg] (1.5667,0)--(1.5667+.2,.6)--(1.5667-.2,.6);
		%		\fill[colr] (5.6667,0)--(5.6667+.2,.6)--(5.6667-.2,.6);
		%		\fill[colb] (5.5867,0)--(5.5867+.2,.6)--(5.5867-.2,.6);
		%		\end{tikzpicture}
		\caption{The $\pvalue_\star(\ygiv)$ \eqref{eq:itmc} values for each of the data sets in Figure~\ref{fig:exWT:1} above, with the same color. The result in gray indicates that the model class~\eqref{eq:WT:ext} is not consistent with the real data according to the \itmc.}\label{fig:exWT:2}
	\end{subfigure}
	\caption{The real data (gray) and synthetic data (colored) together with the \itmc result for the model class $\MC$ in~\eqref{eq:WT:ext}, suggesting that the real data-generating process is not well described by~\eqref{eq:WT:ext}. The synthetic data is generated from models~\eqref{eq:WT:ext} and therefore provide a sanity check for the proposed method.}\label{fig:exWT}\vspace{-1em}
\end{figure}

Our interest in this paper, however, is to answer whether the extended model class~\eqref{eq:WT:ext} describes the true data-generating process well, or if there are more room for modelling improvements. 
We consider a discrete-time formulation of~\eqref{eq:WT:ext}, with the extension to also account for overflow, to be our model class $\MC$ in which the noise is assumed to be Gaussian. All unknown parameters, including the noise variance, are contained in  $\param$.

We evaluate the model with respect to a real data set that contains exogenous input signals. For sake of illustration, we use the same inputs to also generate six synthetic data sets from $\MC$ with physically reasonable values of the unknown parameters. The real and synthetic datasets are shown together in Figure~\ref{fig:exWT:1} and exhibit visually similar characteristics. Intuitively, the proposed model check probes the question: ``is there a statistically meaningful difference between the gray data set (true) and the colored ones (synthetic)''?

The \itmc is implemented using a particle filter to estimate $p(\ysim\mid\param)$ and particle Gibbs with ancestor sampling \citep{LJS:14} to find $\weights(\param\mid\ygiv)$. The result for the real data as well as the synthetic data is found in Figure~\ref{fig:exWT}, which suggests that $\MC$ does not contain a minimum divergence model that is consistent with the observed data. Thus there is scope for improved modelling.

A naive attempt to apply the Ljung-Box method -- which originally is not intended for nonlinear state-space models -- failed to produce any meaningful results. The method recently proposed by \citet{SMW:17} was also applied, which amounts to
\vspace{-.5em}
\begin{enumerate}[(i)]
	\setlength\itemsep{-.3em}
	\item inferring $\weights(\param\mid\ygiv)$,
	\item draw one parameter sample $\param'\sim \weights(\param\mid\ygiv)$,
	\item draw one state trajectory sample from the state smoothing distribution $x_{1:T}\sim p(x_{1:T}\mid\param',\ygiv)$,
	\item analyze whether the corresponding process noise realization follows the Gaussian noise assumption using a z-test.
\end{enumerate} 
\vspace{-.5em}
To implement this, we used again particle filters and particle Gibbs with ancestor sampling. Since this procedure relies on a single random sample from $\weights(\param \mid\ygiv)$, the result of the method varies dramatically each time it is computed, but indicates on average no difference between the real data and the synthetic data, contrary to the result by \itmc. There is no claim that the \itmc and the method by \citet{SMW:17} should be equivalent in any sense. The different results can therefore not be said to be unexpected but they indicate, however, that the \itmc is perhaps more sensitive in finding model inconsistencies with respect to the data.

\section{Conclusions}

We have developed a method to evaluate a specified model class by assessing its capability of reproducing data that is similar to an observed data record. Specifically, we have targeted the minimum divergence models in the class and viewed them as data generators. Using the self-information or surprisal as a test statistic, we formulated the information-theoretic model check (\itmc) as an averaged p-value which can be understood as corresponding to a specific two-sided posterior predictive test. 

We have applied \itmc to both synthetic and real data and obtained promising results, also in comparison to the standard whiteness test and the recent method by \citet{SMW:17}. The usefulness of \itmc for other model classes than dynamic time-series models remains to be investigated, as well as a more thorough theoretical understanding of its behavior. 

\section*{Acknowledgements}
This work has been partly supported by the Swedish Research Council (VR) under contracts 2016-06079 and 621-2014-5874, as well as the Swedish Foundation for Strategic Research (SSF) via the project \emph{ASSEMBLE} (contract number: RIT15-0012).

\bibliography{references} 

\end{document}